# Differentiable Probabilistic Logic Networks

Alexey Potapov[1,2], Anatoly Belikov[1], Vitaly Bogdanov[1], Alexander Scherbatiy[1]

[1]SingularityNET Foundation, The Netherlands
[2]ITMO University, St. Petersburg, Russia
{alexey, abelikov, vitally, alexander.scherbatiy}@singularitynet.io

**Abstract.** Probabilistic logic reasoning is a central component of such cognitive architectures as OpenCog. However, as an integrative architecture, OpenCog facilitates cognitive synergy via hybridization of different inference methods. In this paper, we introduce a differentiable version of Probabilistic Logic networks, which rules operate over tensor truth values in such a way that a chain of reasoning steps constructs a computation graph over tensors that accepts truth values of premises from the knowledge base as input and produces truth values of conclusions as output. This allows for both learning truth values of premises and formulas for rules (specified in a form with trainable weights) by backpropagation combining subsymbolic optimization and symbolic reasoning.

## 1 Introduction

One prominent feature of the OpenCog cognitive architecture [1] is cognitive synergy [2] implying that inference methods of different nature cooperate in such a way that help each other to bypass limitations inherent to each method working independently. In particular, one can consider neuro-symbolic integration [3] or joint usage of probabilistic logical reasoning and evolutionary computations [4] as examples of cognitive synergy in OpenCog.

Cognitive synergy has both conceptual and technical sides. Although there were early attempts to integrate deep neural networks in OpenCog [5], they relied on a specific DNN architecture and didn't integrate DNNs training with OpenCog reasoning. In a companion paper [6], we discuss a more intimate integration of symbolic reasoning with modern DNN frameworks. However, one can consider such integration from a different angle. Instead of focusing on integrating neural networks into symbolic knowledge representations, one can focus on integrating different inference (reasoning/learning) mechanisms.

Indeed, one main inference mechanism in OpenCog is the Unified Rule Engine (URE), which is based on tree search and relies on the Pattern Matcher to find isomorphic subgraphs suitable for rule application. Different forms of reasoning (e.g. fuzzy logic, predicate logic, Bayesian inference) can be implemented using URE. Automatic differentiation can also be implemented by defining certain URE rules.

However, existing deep learning frameworks readily provide a rich functionality for automatic differentiation, error backpropagation and a number of gradient descent optimization techniques with efficient GPU-based implementation. A practical cognitive synergy supposes gaining benefits from utilizing the existing state-of-the-art frameworks built around inference methods of different nature.

In this paper, we implement a number of rule sets for URE including the Probabilistic Logic Networks (PLN) [7] on the base of PyTorch. Specifying PLN truth values as PyTorch tensors and implementing inference formulas as operations over them allows us to make these truth values learnable by gradient descent. Formulas themselves can be made trainable by introducing tensor parameters into them. In this way, we achieve functionality, which implementation on the base of the existing mechanisms in OpenCog can be cumbersome and less efficient.

## 2    Atomspace and Unified Rule Engine

Atomspace is a hypergraph knowledge representation database via which different processes and inference mechanisms interact in OpenCog. Its constituting elements are Atoms, which are generally divided into Nodes and Links (although Links can connect multiple Atoms simultaneously including other Links in contrast to plain graphs). Types of Atoms are introduced based on the needs of representing certain knowledge and reasoning over it. Supplemented by the inference mechanisms, graphs in Atomspace can be considered as programs in a programming language called Atomese.

One basic Node type is `ConceptNode`, which is an atom intended to represent some concept identified by its name. Concepts can participate in a "is-a" relation that is represented by `InheritanceLink` connecting corresponding `ConceptNodes`.

Atomspace may contain such facts that a sparrow is a bird
```
InheritanceLink
  ConceptNode "sparrow
  ConceptNode "bird"
```
and that a bird is an animal
```
InheritanceLink
  ConceptNode "bird"
  ConceptNode "animal",
```
and one might want to infer that a sparrow is an animal.

This can be done with the use of `BindLink`, which connects a query graph to be found in Atomspace, and an implicand graph to be inserted into Atomspace. These graphs contain `VariableNodes` that can be matched against different subgraphs. The Pattern Matcher executing `BindLink`s tries to find all groundings for these variables, for which the query graph appears to be a subgraph of the current knowledge base. For example, executing the following `BindLink`
```
BindLink
  AndLink
    InheritanceLink VariableNode("$X") VariableNode("$Y")
```

```
            InheritanceLink VariableNode("$Y") VariableNode("$Z")
            InheritanceLink VariableNode("$X") VariableNode("$Z")
```
will yield the fact that a sparrow is an animal. The upper-level `AndLink` is not considered by the Pattern Matcher as a Link to be found in Atomspace, but it simply connects clauses – subgraphs that should be found simultaneously with same variable groundings.

Both `ConceptNode`s and `InheritanceLink`s can have associated truth values, which indicate how probable the corresponding Atom and how confident OpenCog about this probability (precise semantics of these truth values depends on the rule set being applied; one of such semantics is defined in the theory of PLN [7]).

How can the truth value be assigned to the implicand subgraph to be inserted into Atomspace? One should have a formula to calculate the resulting truth value from the truth values of clauses in the query. These formulas are implemented in OpenCog by grounded procedures (i.e. procedures written in some external language, e.g. Scheme, Python, etc.), which are accessed through `GroundedSchemaNode`s (which should be wrapped into `ExecutionOutputLink` to be executed with some arguments during pattern matching). The corresponding implicand subgraph in the above `BindLink` can look like

```
        ExecutionOutputLink
          GroundedSchemaNode "scm: deduction-formula"
            InheritanceLink VariableNode("$X") VariableNode("$Z")
            InheritanceLink VariableNode("$X") VariableNode("$Y")
            InheritanceLink VariableNode("$Y") VariableNode("$Z")
```
meaning that there is a function called deduction-formula hand-coded in scheme that should be called with the corresponding arguments. This function should extract truth values from the implicants, calculate the resulting truth value of the implicand, and return the implicand subgraph with the assigned truth value.

Similar rules with corresponding formulas (deduced in the theory of PLN, fuzzy logic or something else) can be defined to infer truth values of compound expressions containing conjunctions, disjunctions, negations, etc. However, several rules might be necessary to chain together to reach a certain conclusion or to calculate a truth value of a compound expression. The Unified Rule Engine is intended to do this sort of chaining in OpenCog either going from premises to conclusions by randomly selecting sources and rules (Forward chainer) or by starting from the target and searching for rules that can produce it and can produce missing premises until all premises can be found in the knowledge base (Backward chainer). The details of URE internals are not trivial and go beyond the scope of the paper. What is important for us here is that the resulting truth values are ultimately calculated by a series of nested calls to grounded functions with some initial truth values as input.

## 3   Tensor Truth Values and Automatic Differentiation

The strong point of the existing design of URE is the possibility to define arbitrary sets of rules as a part of declarative and introspective knowledge base. In particular, one can imagine a set of meta-learning rules for URE, which can help OpenCog to infer new rules. The problem is, however, that truth values are processed by grounded functions which are not transparent for OpenCog and cannot be created or modified by URE (without relying on some other external functions that can do this). Even generally applicable rules with formulas for calculating initial truth values of `ConceptNode`s or `InheritanceLink`s from data are now missing, although Pattern Miner which is now being implemented also with URE can be used in some cases.

For example, if we have a question-answering dataset with questions like "What color is a canary?" accompanied by correct answers, and want to infer the strength of the inheritance links between concepts "yellow", "red", etc. and "color", we will most likely have to write down an ad hoc code for this, while it is commonly desirable to learn everything in end-to-end fashion (while learning to answer questions).

It should be noted that there are `NumberNode` and link types corresponding to arithmetic and comparison operations with `NumberNode`s enabling declarative reasoning over numbers with the use of axioms embodied in URE rules. It might be possible in principle to implement formulas as Atomese programs.

However, truth values are not implemented as Atoms intentionally, because Atoms are unique and immutable (e.g. there can be only one `NumberNode 0.3`, which cannot be changed to `NumberNode 0.5`, which is simply another Atom), searchable, and thus expensive to create, store, and delete (since they can be connected with other Atoms). `NumberNode`s are intended to represent an idea of corresponding numbers. For example, the fact that 7 is a prime number can be stored in the declarative knowledge base, but connecting brightness of all pixels in input video sequence with corresponding `NumberNode`s is inappropriate. Values were introduced to Atomspace for assigning mutable and computationally cheap valuations to Atoms. The necessity to use grounded functions to process Values is thus perfectly natural.

What is missing in OpenCog is some general purpose inference methods over Values. Apparently, the most widely used inference method for subsymbolic real-valued entities is gradient descent, which became conveniently usable due to the frameworks of automatic differentiation on computation or dataflow graphs. To be able to utilize one of these frameworks, one needs to build a computation graph and feeds it with data in a compatible format (e.g. PyTorch or Tensorflow tensors).

Unfortunately, the existing truth values cannot be directly fed there. Moreover, if one tries to implement formulas for URE rules, which convert truth values to tensors, perform some operations over tensors, and convert them back to truth values, chaining these formulas will give a broken computation graph, in which gradients cannot propagate from one part to another.

Because formulas are chained dynamically by URE, we need a framework that allows for dynamic computation graphs. Some frameworks like PyTorch have this feature natively, while others implement it as an additional component (e.g. Tensorflow supported only static graphs initially, and Eager Execution and Fold were added to it later). We utilized PyTorch, although other choices are possible. We implemented an API to access tensor truth values stored as external objects. This API is similar to the existing API for the conventional truth values.

When URE passes these tensor truth values between formulas that perform PyTorch operations over them and set resulting tensors as truth values of implicand graphs, the dynamically constructed computation graphs are not broken and remain suitable for calculating gradients.

## 4     Probabilistic Logic Networks with Differentiable Rules

Probabilistic Logic Networks (PLN) [7] is being implemented as a set of rules for URE. To obtain its differentiable version, one needs to rewrite the formulas of its rules (implemented as `GroundedSchemaNodes` in OpenCog) using PyTorch operations over tensor truth values, that is done here[1]. This enables the error backpropagation from the result of reasoning through the formulas to the truth values of Atoms in the knowledge base. In order to make these values trainable, one needs to initialize them with not just tensors, but with Torch variables with set requires_grad flag.

Let us explain this on a simple example of the classical modus ponens inference rule, which is not only a crisp rule in the form "$A{\rightarrow}B, A \vdash B$", but also has a formula to calculate the truth value of the conclusion depending on the truth values of premises in PLN. The precise formula should look like:

$P(B) = P(B|A)P(A) + P(B|\neg A)P(\neg A),$     (1)

where $P$ is the strength of the truth value that corresponds to probability (we don't consider the confidence component of truth values here).

Let us imagine that we ask PLN to infer if a particular apple or banana is yellow, red, or green. It can be done in the context of visual question answering, but we will not consider the use of visual cues here.

Thus, we have a set of facts, which look like
```
    EvaluationLink
      PredicateNode "apple"
      ConceptNode "apple-001"
```
with certain tensor truth values (1.0, 1.0).

Then, we have a set of general facts saying that every apple is green or red, e.g.
```
    ImplicationLink
      LambdaLink
        VariableList
          TypedVariableLink
            VariableNode "$X"
            TypeNode "ConceptNode"
          EvaluationLink
            PredicateNode "apple"
            VariableNode "$X"
      LambdaLink
        VariableList
          TypedVariableLink
```

---

[1] https://github.com/singnet/opencog/tree/master/opencog/torchpln

```
            VariableNode "$X"
            TypeNode "ConceptNode"
        EvaluationLink
            PredicateNode "green"
            VariableNode "$X"
```
or simply
```
    ImplicationLink
      PredicateNode "apple"
      PredicateNode "green"
```
Each training query makes PLN to apply modus ponens rule to infer the truth value of the conclusion:
```
    EvaluationLink
      PredicateNode "green"
      ConceptNode "apple-001"
```
The inferred truth value can be compared to the ground-truth fact: in our experiments, we sample the color of each fruit with predefined probabilities, which are unknown to OpenCog.

To learn the truth value of these implication links (connecting each type of fruit with each color) one needs to specify a loss function and use an optimizer to minimize it. Cross-entropy is a natural choice since the result of PLN reasoning is truth value, which strength component corresponds to probability.

Let us consider a subset for a particular fruit and particular color. Then, the corresponding loss function will be

$$Loss = \sum_i \left[ y_i \log(p_i) + (1 - y_i) \log(1 - p_i) \right] \quad (2)$$

where $y_i=1$ if $i$-th instance of this fruit is of this color (e.g. apple-001 is green), and $y_i=0$ otherwise. $p_i$ is the inferred probability that this fruit is of this color, i.e. $P(B)$ in (1). Trivially, if $P(A)=1$ in (1) since we set it for all such links as
```
    EvaluationLink
      PredicateNode "apple"
      ConceptNode "apple-001",
```
then $P(B)$ as the probability for
```
    EvaluationLink
      PredicateNode "green"
      ConceptNode "apple-001"
```
will simply be equal to $P(B|A)=P(green|apple)$, i.e. the truth value of the corresponding implication link. Thus, we can rewrite (2) as

$$Loss = \sum_i \left[ y_i \log P(B|A) + (1 - y_i) \log(1 - P(B|A)) \right] = n_1 \log P(B|A) + n_0 (1 - \log P(B|A)),$$

where $n_0$ is the number of not green apples, and $n_1$ is the number of green apples in the training set. Minimum of this loss function can be found from:

$$0 = \left[ n_1 \log P + n_0 (1 - \log P) \right]'_P = n_1/P - n_0/(1-P) \Rightarrow P(B|A) = \frac{n_1}{n_0 + n_1}$$

And indeed, this result is obtained by gradient descent applied to the loss function (2) that penalizes the reasoning for inferring values differing from ground truth. That is, gradient descent finds the optimal truth value of the implication link $A(x){\rightarrow}B(x)$ as a fraction of examples $N[A(x)$ is true and $B(x)$ is true$]/N[A(x)$ is true$]$, e.g. the fraction of green, but not red or yellow, apples. This is a desirable result. It should be noted that we don't use an explicit equation here, but only penalize PLN for wrong conclusions.

Truth values are usually modified by URE rules themselves, and the considered simple task can be handled by PLN with the use of direct estimation of the frequencies. Do we need to use additional tools like gradient descent to modify them?

However, there are some complications to handle more difficult cases entirely in Atomese. First of all, the update rules should be applied to the inference trees, which can contain much more steps than the application of a single rule. In such cases, we would prefer not to manually infer explicit equations for optimal truth values, but to obtain them automatically.

These traces can be made transparent for OpenCog (i.e. can be represented as Atomspace subgraphs, which is done in the inference control experiments [8]). However, the error should be back-propagated through not just URE rules, but through formulas attached to them, and these formulas are not transparent. Thus, either these formulas should be implemented in Atomese (that will have some drawbacks), or each grounded formula should be accompanied by a hand-coded backward formula. Our approach avoids these subtleties, and also enables trainable rules for URE.

## 5    Trainable Rules

In order to obtain a trainable rule, one needs just to parameterize a grounded formula implemented as a torch computation graph. Parameters specified as torch variables with set requires_grad flag will be automatically optimized w.r.t. the specified loss function. In the course of this optimization, one can hope to obtain correct formulas for patterns of reasoning defined by rules. Since the Atomese rule induction itself can be approached with the use of OpenCog modules (e.g. MOSES, Pattern Miner), learning grounded formulas by gradient descent can complement these symbolic mechanisms to achieve an end-to-end rule induction. However, we will consider the case of optimizing a formula for a defined rule here.

The first question, which arises while attempting to introduce trainable formulas, is in what form to specify them. A straightforward way to represent such formulas as (1) as members of some parametric family would be to insert multiplicative and additive trainable variables into them. Unfortunately, this will not lead to a desirable result, because nothing will prevent gradient descent from setting such values for the parameters that the formula application will lead to degrade values of resulting probabilities outside [0, 1] range. Trainable truth values will also go outside this range as a result.

Instead of probabilities, one can work with some other quantity with the unrestricted domain that maps to [0, 1] by some transformation (e.g. sigmoid function).

However, one will also need to perform computations in the space of these quantities (and simple equations as (1) in this space will have more complex form) and to map the final result into probability. Thus, one can simply approximate the formulas that take probabilities as input and produce them as output with a parametric family of functions with restricted range. For example, one can try to approximate (1) by

$$P(B) \approx \sigma(w_0 P(A)P(B|A) + w_1 P(A) + w_2 P(B|A) + w_3) \qquad (3)$$

If we use ground-truth values for $P(B|A)$ together with fixed values for $P(A)$, it is possible to learn such weights in (3) by gradient descent that will give a good approximation of (1) despite of additional non-linearity in (3). Thus, a good approximation of the formula for the modus ponens rule will be learned from examples (the code can be found here[2]). Of course, to be able to take all input probabilities into account correctly, it should be provided by different combinations of their values in the training set.

One can notice that (3) looks like one layer of a neural network (although with some additional input "features" like $P(B|A)P(A)$). Apparently, one can use generic DNNs to approximate arbitrary formulas for different rules.

However, a problem arises if we try to learn both the formula and unknown truth values simultaneously. The model will have too many degrees of freedom to learn some transformed quantities instead of $P(B|A)$ and to account for this transformation in the formula. Although the final results of inference will be as precise as before, the inferred truth values of implication links will have no clear semantics. This problem can be bypassed by training the model on a richer dataset with different input probabilities provided as ground truth in different examples, so the formula will have to be consistent with different use cases, and thus the inferred truth values in one set of examples will be consistent with ground truth values in another set of examples. However, a more principled approach to ensure the probabilistic semantics of truth values learned together with formulas can be desirable.

## 6   Conclusion

We have introduced a differentiable version of Probabilistic Logic Networks that is achieved by rewriting the formulas for its rules with the use of operations over truth values represented as PyTorch tensors. Chaining of PLN rules by the Unified Rule Engine yields an unbroken computation graph over tensors that accepts truth values of premises from the knowledge base as input and produces truth values of conclusions as output. Automatic differentiation and gradient descent based optimization methods of PyTorch are thus readily applicable to this graph. Optimization is performed over tensors specified as PyTorch variables.

We have conducted a basic experiment with the modus ponens rule, and have shown that an appropriate truth value of implication is learned by backward propagation of errors from the conclusions made by PLN to the truth values of premises.

---

[2] https://github.com/singnet/opencog/tree/master/examples/python/tensortruthvalue

If PyTorch variables are introduced into formulas of rules, these formulas themselves become trainable. While learning precise formulas is not straightforward, good approximations to them can easily be learned in cases when truth values of all premises and conclusions are known.

Formulas can be learned together with truth values of implication links. However, some subtle issues arise here. While such trainable PLN learns to produce the desirable conclusions, truth values can be learned as non-linearly transformed desirable values with formulas accounting for this transformation. Training PLN formulas with examples, some of them have known truth values for one kind of premises, while others have them for other premises, can help. However, a more principled solution that forces preserving the semantics of truth values might be desirable.

## References

1. Goertzel, B.: Opencog prime: A Cognitive Synergy based Architecture for Embodied Artificial General Intelligence. In: ICCI 2009, Hong Kong (2009)
2. Goertzel, B.: Toward a Formal Model of Cognitive Synergy. T. Everitt, B. Goertzel, A. Potapov (Eds.): AGI 2017, Springer: Lecture Notes in Artificial Intelligence, vol. 10414, pp. 13–22 (2017)
3. Goertzel, B.: Imprecise Probability as a Linking Mechanism Between Deep Learning, Symbolic Cognition and Local Feature Detection in Vision Processing. AGI 2011, Springer: Lecture Notes in Artificial Intelligence, vol. 6830, pp. 346-350 (2011)
4. Goertzel, B., et al.: Speculative Scientific Inference via Synergetic Combination of Probabilistic Logic and Evolutionary Pattern Recognition. J. Bieger, B. Goertzel, A. Potapov (Eds.): AGI 2015, Springer: Lecture Notes in Artificial Intelligence, vol. 9205, pp. 80–89 (2015)
5. Goertzel, B., Sanders, T., O'Neill, J.: Integrating Deep Learning Based Perception with Probabilistic Logic via Frequent Pattern Mining. K.-U. Kühnberger, S. Rudolph, P. Wang (Eds.): AGI 2013, Springer: Lecture Notes in Artificial Intelligence, vol. 7999, pp. 40–49 (2013)
6. Potapov, A., Belikov, A., Bogdanov, V., Scherbatiy, A.: Cognitive Module Networks for Grounded Reasoning. P. Hammer et al. (Eds.): AGI 2019, LNAI 11654 (2019) https://doi.org/10.1007/978-3-030-27005-6_15
7. Goertzel, B., Iklé, M., Goertzel, I.L.F., Heljakka, A.: Probabilistic Logic Networks: A Comprehensive Conceptual, Mathematical and Computational Framework for Uncertain Inference. Springer. p. 333 (2008)
8. Geisweiller, N.: Partial Operator Induction with Beta Distributions. AGI 2018, Springer: Lecture Notes in Computer Science, vol. 10999, pp. 52–61 (2018)